\documentclass{Interspeech}

\interspeechcameraready

\title{Prosodic Structure Beyond Lexical Content: A Study of Self-Supervised Learning}

\author[]{Sarenne}{Wallbridge*}
\author[]{Christoph}{Minixhofer*}
\author[]{Catherine}{Lai}
\author[]{Peter}{Bell}

\affiliation{Centre for Speech Technology Research}{University of Edinburgh}{UK}
\email{\{s.wallbridge, christoph.minixhofer, peter.bell, c.lai\}@ed.ac.uk}
\keywords{prosody, self-supervised learning, emotion recognition, prominence prediction, phonetic segmentation}
\interspeechcameraready

\usepackage{xcolor}
\usepackage{comment}
\usepackage{hyperref}
\usepackage{multirow}
\usepackage{cleveref}
\usepackage{booktabs}
\usepackage{makecell}
\usepackage{siunitx}
\usepackage{subcaption}
\usepackage{graphicx}
\usepackage{subcaption}
\newcommand{\burn}{\textsc{Burnc}}
\newcommand{\rav}{\textsc{Ravdess}}
\newcommand{\timit}{\textsc{Timit}}

\newcommand\blfootnote[1]{%
  \begingroup
  \renewcommand\thefootnote{}\footnote{#1}%
  \addtocounter{footnote}{-1}%
  \endgroup
}

\let\OLDthebibliography\thebibliography
\renewcommand\thebibliography[1]{
  \OLDthebibliography{#1}
  \setlength{\parskip}{2.1pt}
  \setlength{\itemsep}{0pt plus 0.1ex}
}

\begin{document}

\maketitle

\begin{abstract}

People exploit the predictability of lexical structures during text comprehension. 
Though predictable structure is also present in speech, the degree to which prosody--e.g., intonation, tempo, and loudness--contributes to such structure independently of the lexical content is unclear.
This study leverages self-supervised learning (SSL) to examine the temporal granularity of structures in the acoustic correlates of prosody. 
Representations from our proposed Masked Prosody Model can predict perceptual labels dependent on local information, such as word boundaries, but provide the most value for labels involving longer-term structures, like emotion recognition.
Probing experiments across various perceptual labels show strong relative gains over untransformed pitch, energy and voice activity features.
Our results reveal the importance of SSL training objective timescale and highlight the value of complex SSL-encoded structures compared to more constrained classical structures.

\end{abstract}

\section{Introduction}
\blfootnote{* for joint authorship.}
Cognitive science theories often describe human language comprehension as a process of leveraging the predictable structure of linguistic signals, i.e., their \emph{systematicity}, to generate expectations about the upcoming signal \cite{hale2001probabilistic}.
Though empirical support stems primarily from written language comprehension, 
listeners also exploit systematicity to predict features of upcoming speech, e.g., informational content \cite{brown2015metrical, 
wallbridge2023}, the position of prosodic focus \cite{ip2017intonation}, and the length of upcoming units (sentences \cite{grosjean1983howlong} and conversational turns \cite{Ekstedt2022HowMD}). 
However, the extent to which this predictive capacity is contingent on the structure of lexical information or of its acoustic realisation is unclear.
To better understand how listeners generate expectations about upcoming speech, we investigate systematicity in non-lexical features of prosody---relative pitch (F0) and loudness (energy)\footnote{In this paper, we now use `prosody' to refer to non-lexical aspects of speech including relative pitch (F0), energy, and timing features.}.

We address two challenges in studying the structure of prosody.
First,  prosody operates at multiple temporal scales, making it difficult to define a 
task-agnostic unit (cf. \cite{Cutler2014HowPI}).
Second, 
prosody functions \textit{in conjunction} with lexical content---e.g.,
interpretations of prosodic changes to discourse markers are restricted by lexical semantics \cite{lai2010you}.
Still, prosody also conveys information \textit{independently} of lexical content, indicating that its acoustic correlates exhibit systematicity. 
For example, people
can be primed to different syntactic disambiguation strategies by prosodic differences in de-lexicalised audio \cite{mills20delexicalised}.
Moreover, pitch contours can help classify reported speech \cite{cervone2015towards}. 
To address both challenges, we apply Self-Supervised Learning (SSL), a mechanism that exploits complex structures across timescales, to study the acoustic correlates of prosody.


We introduce a novel Masked Prosody Model (MPM) which encodes correlates of pitch, loudness and voice activity by learning to reconstruct corrupted feature sequences. 
By altering the corruption strategy, 
we ask 
1) \textit{Can SSL capture predictable structure in prosody, independent of lexical content?} and
2) \textit{How does corruption timescale affect the utility of representations for predicting perceptual speech labels?}
We select labels that rely on structures at different temporal granularities. Using linear probes,
we find a task-dependent effect of corruption granularity on 
the utility of resulting representations and propose a masking strategy for generalisable representations.
Comparing MPM to a constrained prosodic encoding reveals that the complex contextualisation of SSL is particularly useful for predicting abstract perceptual labels. Stronger probes illustrate how SSL and
task-specific structures interact, underscoring the complexity of prosodic structures. 
Finally, comparing MPM to SSL representations of the full speech signal highlights how prosodic structure contributes to different tasks.


\section{Background}
\subsection{Self-supervised learning}
SSL mechanisms encode the structures of their training data by learning to reverse a corruption function---e.g., removing future, past, or intermediate context. 
Without a need for external labels, SSL can exploit large amounts of unlabeled data. 
The resulting representations are useful for downstream tasks with limited labelled data, either as input to task-specific models or through fine-tuning \cite{donahue2014decaf}.
Transformers trained with SSL capture both global and local structures and have been successfully applied to text \cite{devlin2019bert}, speech and a host of other signals \cite{ofer2021language}; 
popular models of speech include wav2vec \cite{schneider2019wav2vec} and HuBERT \cite{hsu2021hubert}.
However, it is unclear which input structures SSL exploits during training. In particular, it is unclear how much prosodic structure is maintained in SSL representations of speech  \cite{deseyssel2023prosaudit}. 

\subsection{Representations of prosody}



Neural representations of prosody have recently gained popularity, often 
with the aim of controlling prosodic realisations of text 
inputs. 
Many of these encodings use a subtractive definition of prosody: the variation remaining once phonetic, speaker, and channel information are removed \cite{SkerryRyan2018Towards}. 
Disentanglement of these factors has been induced through carefully-tuned bottlenecks or training data design \cite{qu2023disentangling,ioannides23paralinguistic}. 
While such representations are useful for speech generation and tasks like emotion recognition, the degree of achieved (and achievable) disentanglement is unclear as they show sensitivity to e.g., speaker and lexical perturbation
\cite{SkerryRyan2018Towards, sigurgeirsson2023transfer}.
As such, studying prosody isolated from lexical content with these representations is difficult.

Representations based instead on acoustic correlates of prosody are independent of lexical information but less well-studied.
One of the few widely used representations of this type is Continuous Wavelet Transform (CWT) across F0, energy, and duration features for automatic annotation of local prosodic events \cite{Suni2017HierarchicalRA}.
CWT captures hierarchical structure by convolving the original signal with sets of wavelets. 
Using a rule-based labelling method, 
CWT features achieve comparable performance to supervised methods for word-level prominence and boundary detection and are useful for generating prosodic renditions that are more faithful to references \cite{Suni2020ProsodicPA}. These results support the importance of hierarchical structure in prosody, which has long been highlighted by linguistic theories \cite{Selkirk1984PhonologyAS, Beckman1986StressAN}.
Other work has used the predictability of word-level pitch and intensity features to detect prominent words \cite{kakouros2015automatic}. 
Learned representations of F0 and delexicalised speech have also proven useful for 
classifying persuasion and sarcasm \cite{weston2021learning}. These works demonstrate that prosodic correlates exhibit systematicity but leave open the question of at what time scale.

\section{Experiments}


Inspired by masked language models for text, we introduce a Masked Prosody Model (MPM) that learns to reconstruct corrupted sequences of pitch, loudness and voice activity during pretraining \cite{devlin2019bert}.
We investigate the effects of different corruption strategies on the utility of resulting representations across a set of downstream tasks that depend on prosodic features at varying timescales.  
While text units are much more intuitive to define than speech units, mask size
plays an important role in the quality of text representations \cite{levine2020pmi}.
Similarly, we expect temporal and feature granularity to impact the complexity and value of MPM training.
We use linear probes to compare MPM representations to hierarchical CWT encodings of acoustic correlates of prosody. Additionally, we use Conformer probes to compare the value of generic and task-dependent structures in both prosodic correlates and the full speech signal.

\subsection{Masked Prosody Model}
\label{sec:mpm} 

\begin{figure}
    \centering
    \includegraphics[width=.7\linewidth]{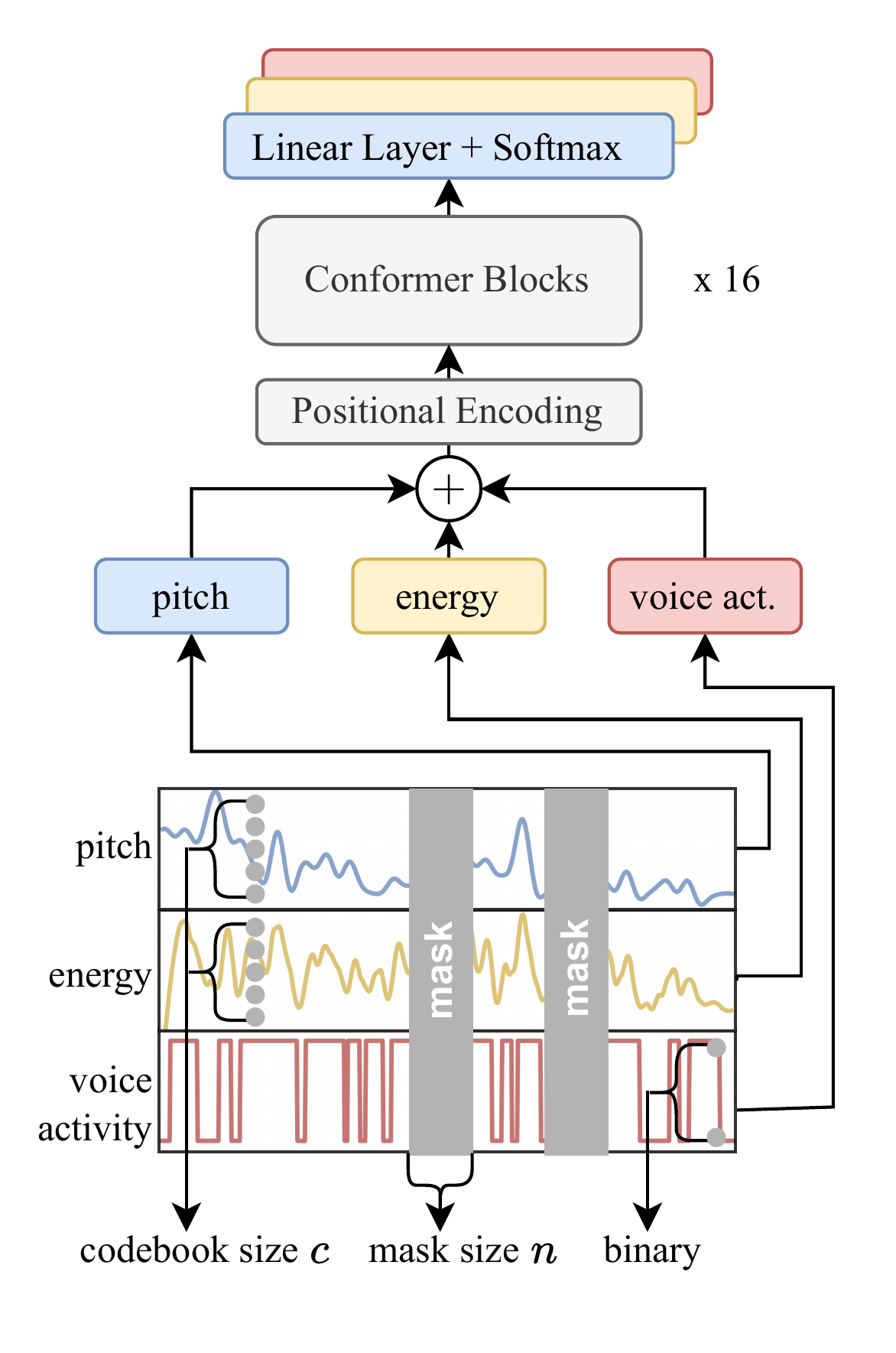}
    \vspace{-12pt}
    \caption{Architecture of the Masked Prosody Model.}
    \label{fig:prosody_model}
    \vspace{-10pt}
\end{figure}

The architecture of the Masked Prosody Model, shown in \Cref{fig:prosody_model}, consists of 
separate input-output sequences for each of the prosodic features and Conformer blocks \cite{Gulati20conformer} which are well-suited to their high resolution and continuous nature.
Pitch (F0) and voice activity are extracted using the WORLD vocoder \cite{morise2016world}. Energy is computed as the Root-Mean-Square (RMS) of each Mel Spectrogram frame. All features are extracted at a resolution of $\approx$10ms
and normalised across the utterance to produce feature contours capturing important perceptual cues \cite{rosenberg2010classification, kimball2016pitch}. Though normalisation removes information about e.g., the relative feature values for individual speakers, it allows encoding of unseen speakers.
Before masking, input sequences are quantized into respective codebooks $P$, $E$ and $V$ of size $c$. Random segments of the aligned sequences, each with a mask length $m$, are masked until $(50\pm5)\%$ of the input signal remains. The model is trained to reconstruct each feature with an independent Categorical Cross Entropy loss. In training, feature losses are normalized by $1/\log{c}$ and their sum is optimized.\footnote{
We release the code and pretrained weights for the MPM model at \href{https://github.com/minixc/masked_prosody_model}{\texttt{github.com/minixc/masked\_prosody\_model}}}


To evaluate model representations, features from the 8th MPM Conformer layer are extracted as intermediary network representations are robust across tasks \cite{yosinski2014intermediate}. When aggregating at word or utterance levels, we compute the mean and maximum features for the target unit and concatenate them. Representations are evaluated using both linear probing \cite{alain2016understanding} by training a linear classifier, and a more powerful Conformer classification model with 2 layers.

\subsection{Downstream Tasks}

\begin{table*}[t]
\vspace{-8pt}
\caption{Probe evaluations using prosody representations (MPM, CWT, and untransformed representations of pitch, energy, and VAD) and speech representations (mel spectrogram, wav2vec, and HuBERT) as input. \timit\ syllable segmentation is evaluated with syllable error rate (SeR) and the correlation between true and predicted number of syllables; \burn\ tasks with F1; \rav\ emotion classification with weighted (WA) and unweighted accuracy (UA). Best task performance for prosody and speech models are in \textbf{bold}.}
\vspace{-8pt}
\begin{subtable}{\textwidth}
\caption{Downstream task performance with the \textbf{linear probe}.}
\centering
\begin{tabular}{@{}lll|cc|cc|cc|cc}
\toprule
\multicolumn{3}{c|}{\textbf{Input Feature}} & \multicolumn{2}{c|}{\textbf{\timit}} & \multicolumn{2}{c|}{\textbf{\burn} (F1)} & \multicolumn{2}{c}{\textbf{\rav}} \\
\multicolumn{1}{c}{\textbf{Type}} &\multicolumn{1}{c}{\textbf{Name}} & \textbf{Mask} & \textbf{SER$\downarrow$} & \textbf{Corr.$\uparrow$} & \textbf{Boundary} & \textbf{Prominence} & \textbf{WA} & \textbf{UA} \\ 
\midrule
\multirow{3}{*}{Speech} & HuBERT           & -- & 12.9           & \textbf{0.90}   & \textbf{0.32}   & \textbf{0.59}   & \textbf{0.63}    & \textbf{0.65} \\ 
                        & Wav2Vec          & -- & \textbf{12.8}  & 0.85            & 0.28            & 0.56            & 0.54             & 0.55  \\
                        & Mel Spectrogram  & -- & 18.3           & 0.75            & 0.18            & 0.54            & 0.23             & 0.21  \\
\midrule
\multirow{6}{*}{Prosody} & \multirow{4}{*}{Masked Prosody Model} & 4      & \textbf{14.9} & 0.85          & 0.27          & 0.51          & 0.15          & 0.16          \\
                                   &  & 16     & 15.3          & \textbf{0.86} & \textbf{0.28} & \textbf{0.58} & 0.17          & 0.17          \\
                                   &  & 128    & 16.0          & 0.78          & 0.25          & 0.50          & 0.21          & 0.22          \\
                                   &  & random & 15.7          & 0.81          & 0.27          & 0.57          & \textbf{0.24} & \textbf{0.23} \\   
\cmidrule(l){2-9} 
                        & Pitch, Energy, VAD (CWT)              & --     & 19.8          & 0.71          & 0.02          & 0.39          & 0.20          & 0.19         \\
                        & Pitch, Energy, VAD                    & --     & 23.3          & 0.62          & 0.07          & 0.49          & 0.10          & 0.09          \\
\bottomrule
\end{tabular}
\label{tab:linear_eval}
\end{subtable}

\begin{subtable}{\textwidth}
\caption{Downstream task performance with the \textbf{conformer probe}.}
\centering
\begin{tabular}{@{}ll|cc|cc|cc|cc}
\toprule
\multicolumn{2}{c|}{\textbf{Input Feature}} & \multicolumn{2}{c|}{\textbf{\timit}} & \multicolumn{2}{c|}{\textbf{\burn} (F1)} & \multicolumn{2}{c}{\textbf{\rav}} \\
\multicolumn{1}{c}{\textbf{Type}} & \multicolumn{1}{c|}{\textbf{Name}} & \textbf{SER$\downarrow$}    & \textbf{Corr.$\uparrow$} & \textbf{Boundary} & \textbf{Prominence} & \textbf{WA} & \textbf{UA} \\ 
\midrule
\multirow{3}{*}{Speech} & HuBERT          & \textbf{8.7} & \textbf{0.96} & \textbf{0.52} & \textbf{0.61} & \textbf{0.65} & \textbf{0.63} \\ 
                        & Wav2Vec         & 10.2         & 0.95          & 0.51          & 0.60          & 0.63          & 0.63          \\
                        & Mel Spectrogram & 16.0         & 0.79          & 0.49          & 0.61          & 0.42          & 0.43          \\
\midrule
\multirow{3}{*}{Prosody} & Masked Prosody Model (random) & \textbf{14.1} & \textbf{0.82} & \textbf{0.53} & \textbf{0.65} & \textbf{0.37} & \textbf{0.36} \\
                         & Pitch, Energy, VAD (CWT)    & 16.1          & 0.78          & 0.46          & 0.51          & 0.31          & 0.29          \\
                         & Pitch, Energy, VAD          & 17.0          & 0.76          & 0.44          & 0.50          & 0.22          & 0.21          \\
\bottomrule
\end{tabular}
\label{tab:conformer_eval}
\end{subtable}
\vspace{-8pt}
\end{table*}



We investigate prosodic structure by predicting
human annotations known to be at least partially driven by prosodic features at different temporal scales. 
First, we assess whether representations capture local structure using the task of \textbf{syllable segmentation}. 
Syllabic units are often defined by changes to acoustic features including intensity and pitch \cite{parker2002quantifying} and are fundamental to models of speech perception; for example, infants without prior linguistic knowledge and adults in artificial language learning experiments are thought to rely on syllable-level units to acquire language \cite{jusczyk1995young,matzinger2021influence}. 
Second, prosody is known to support the linguistic organisation of speech above words through the demarcation of structure and salience 
\cite{cole2015prosody, deseyssel2023prosaudit}. As such, we test \textbf{prominence and break detection} where CWT features have already proved useful.
Finally, we investigate
utterance-level systematicity using \textbf{emotion classification}, as affective information can be conveyed through prosody at the utterance level \cite{batliner2010segmenting}.
Rather than absolute performance, we are interested in the \textit{relative} predictive power of different representations.


We evaluate \textbf{syllable segmentation} using {\textbf{\timit}}, a set of high-quality recordings from 630 speakers 
each reading ten phonetically-rich sentences \cite{garofolo1993timit}.  For performing syllabification on {\textbf{\timit}}, previous works \cite{jiao2015convex,yuan2021automaticrecognitionsuprasegmentalsspeech} have framed the task as predicting vowel positions as a proxy for the number of syllables. Evaluation is done using Speaking Rate Error Rate (SER), which is defined as the absolute difference between actual and predicted syllables, divided by the actual number of syllables. The second metric is the correlation coefficient (Corr.) between the actual and predicted numbers of syllables per utterance.
Given that syllables are marked by local rhythmic modulations in intensity and pitch, we don't expect MPMs with wide masks to provide much benefit over untransformed input features \cite{rasanen2018pre,bagou2002contribution}.
\textbf{Prominence and break detection} are evaluated on \textbf{\burn} (the Lab News portion of Boston University Radio News Corpus) 
\cite{ostendorf1995boston}. In this corpus, prominence is annotated with ToBI pitch accent types and boundaries between words are labelled for strength on a scale $\{0,..,4\}$ \cite{silverman1992TOBI}. We follow previous works and convert both tasks to binary classification where a word is prominent if any constituent syllable carries an accent (\{H*, L*, L*+H, L+H*, H+, !H*\}) and boundaries consist of the strongest break indexes (\{3,4\}) \cite{Suni2017HierarchicalRA}. 


There is no established train and test split for \textbf{\burn}, and previous works use differing filtering and data selection methodologies \cite{ananthakrishnan2007automatic,Suni2017HierarchicalRA}. Due to this and the small dataset size, we use 5-fold cross validation and use the full dataset, including noisy samples.
Using a handcrafted classification method over CWT features of pitch and energy, \cite{Suni2017HierarchicalRA} report F1 respective performances of 0.85 and 0.59 for prominence and boundary detection. Pitch, energy, and durational information are useful, particularly in combination, for both tasks \cite{ludusan2014towards,kalinli2007saliency}. We expect the additional flexibility of MPM encoding to be beneficial.
For \textbf{emotion classification}, we use the \textbf{\rav} dataset (Ryerson Audio-Visual Database of Emotional Speech and Song Dataset), 1,440 utterances spoken by 24 speakers in eight emotions \cite{livingstone2018ryerson}. 
We select \rav\ as emotion labels were validated with a wide battery of listening tests. 
Though we know of no purely prosodic baselines on this dataset, handcrafted prosodic features \cite{cao2014prosodic} and neural representations of delexicalised speech \cite{weston2021learning} have proved useful for emotion classification. 
This dataset consists of emotional renditions of two neutral phases.
Though lexical content doesn't affect our proposed representations, this feature enables interesting comparisons with representations of the full speech signal.

\subsection{Experimental Setup}

We define MPM corruption strategies as 
combinations of mask $m$ and codebook $c$ sizes and expect the utility of self-supervision to be bounded by these parameters; e.g., small masks may be predicted trivially from continuous acoustic features, while recovering large masks will become impossible.
In preliminary experiments, $m\in\{1,2,4,8,16,32,64,128\}$ and $c\in\{4,8,16,32,64,128,512\}$ were tested;
the smallest mask sizes consistently performed poorly and the best performance for all tasks was achieved with $c=128$. For brevity, we report on strategies with $c=128$ and a representative range $m \in \{4,16,128\}$.
Motivated by \cite{joshi2020spanbert} who find that masking randomly-sized spans of text yields more useful representations than masking individual tokens, we propose \textit{random masking} as a generic corruption strategy; this involves sampling $m$ uniformly between $(1,128)$ for each batch. 
Using each corruption strategy, we train an MPM on the full LibriTTS dataset of over 500 hours of audiobook data \cite{Zen2019LibriTTSAC}. Each model is trained for 10k steps ($\approx$8 epochs, 30 minutes of TPUv3 runtime) with a batch size of 256, over utterances truncated to a maximum length of 6 seconds. 

To enable comparison, CWT representations are generated from the same input features as MPMs (see \Cref{sec:mpm}) and no hyperparameter tuning or architecture adaptations are performed for downstream tasks. 
For topline results including the full speech signal, we use Wav2Vec Base \cite{schneider2019wav2vec}, HuBERT Base \cite{hsu2021hubert} and Mel Spectrograms. For Wav2Vec and HuBERT, we use the model checkpoints trained on \(\approx\)900h of LibriSpeech, which is comparable to the amount of data seen by the MPM.

The linear and Conformer (2 Conformer blocks, 5M parameters) classification probes are trained for 1000 steps with a batch size of \num{32}, the AdamW optimizer, and a linear learning rate schedule with a maximum learning rate of \num{4e-5} and 100 warmup steps.








\vspace{-1em}
\section{Results}
\subsection{Effect of Corruption Strategy}
\label{sec:model_selection}

In the MPM section of \Cref{tab:linear_eval}, we use a linear probe to directly compare the performance achieved from models with $c=128$ and $m \in \{4,16,128\}$ across the downstream tasks.
We find that syllable segmentation benefits from smaller mask sizes; larger mask sizes may not encode sufficiently fine-grained local structures. Though boundary detection shows less sensitivity to mask size than prominence detection, the midsized mask performs best for both. Emotion recognition improves steadily with mask size, suggesting that some relevant features are only present when large mask sizes are encountered during training.
The random-mask model performs well across all tasks. Although some local information useful for boundary detection and syllable segmentation may be degraded, it outperforms all corruption strategies for emotion recognition, indicating that random masking encodes both global and local structures and that emotion recognition benefits from this combination.

\subsection{Comparing Structural Encodings of Prosody}


We investigate the value of structure at different timescales by comparing MPM random-mask representations to the untransformed input features and CWT hierarchical representations.

Using \textbf{linear probes}, \Cref{tab:linear_eval} shows the random-mask MPM consistently outperforms the untransformed and CWT features; the greatest relative gains of MPM are for \burn\ boundary detection. The CWT prosodic features do not outperform the untransformed features for the \burn\ tasks. This could indicate that some of the local features needed for both tasks are obscured by this transformation.
Although both CWT and MPM outperform the untransformed input features for emotion recognition, MPM representations provide additional predictive power. This suggests that while hierarchical information is useful for this task, the additional structures learned through SSL are more valuable.
MPM only slightly outperforms the input features for \timit\ syllable segmentation, confirming our expectation that this task doesn't require wide contextualization.
For tasks that are expected to involve longer-ranging temporal context---prominence detection in \burn\ and the \rav\  emotion classification---the random mask model performs best overall.

In the Prosody section of \Cref{tab:conformer_eval}, we use more powerful \textbf{Conformer probes} to test whether the self-supervised MPM encodings provide value over learnable task-specific structures.
Compared to the linear probe, the added capacity of the conformer produces higher performances across all representations; however, the magnitude of gains varies between tasks. \timit\ syllable segmentation detection shows relatively small improvements, indicating that the capacity of the probe was not a bottleneck on MPM performance. Boundary and prominence detection and emotion classification show larger improvements, suggesting that additional task-specific structures can be learnt by the Conformer probe. 
Though both CWT and MPM structures outperform the untransformed input for \rav\ emotion classification, the flexibility of MPM is more valuable. The hierarchical structures encoded by CWT may not be sufficiently expressive for this more abstract task.
Interestingly, though there are differences between their performance using the linear probe, the untransformed input features and CWT encodings achieve comparable performance for syllable segmentation and boundary and prominence detection with the Conformer probe. The hierarchical structures encoded by CWT 
only offer slight benefits over the task-specific structures learned from untransformed input features. MPM representations outperform both representations, again suggesting that useful information is more accessible in MPM features.

\subsection{Investigating Additional Information in Speech}
We also compare the SSL representations of the acoustic correlates of prosody to those of the full speech signal in \Cref{tab:conformer_eval}. Doing so highlights what information may be useful for particular downstream tasks.  
We might expect speech representations to consistently outperform MPM which only has access to a subset of information in the speech signal; however, \Cref{tab:conformer_eval} shows that MPM offers slight improvements over both wav2vec and HuBERT for \burn\ boundary and prominence detection using the Conformer probe. This could indicate that salient information is more accessible in MPM representations.
Although the SSL speech representations achieve the best \timit\ syllable segmentation performance, MPM features outperform the mel spectrogram: though phonetic information
is valuable for such segmentation, this result highlights that
syllable structure in \timit\ is conveyed through multiple speech features.
All representations of the full speech signal outperform the prosody-only representations for emotion classification. 


\rav\ utterances are realisations of the same underlying lexical content \cite{tian2016recognizing}, and therefore this cannot be explained by additional lexical content present in the full speech data. \rav\ performance therefore depends on other features of speech beyond pitch, intensity, and voice activity, which could be explored in future work.

\section{Discussion \& Conclusions}
We find that self-supervised methods can produce useful representations of the acoustic correlates of prosody, indicating that prosody exhibits predictable structure---systematicity---independently of lexical content.
Our comparisons of corruption strategies reveal structure across timescales: 
syllable segmentation benefits from small masks while larger masks produce effective representations for detecting phrasal boundaries and emotion. 
By comparing MPM to hierarchical CWT representations of prosody, we find evidence for additional structural complexity; the flexibility of SSL is particularly valuable for emotion classification which involves complex dependencies across timescales. 
Our random-mask strategy encodes both long- and short-term structures, offering a generic representation of prosody that performs well across all tested tasks. To the best of our knowledge, we are the first to quantify the value of encoding structure of prosodic correlates isolated from lexical/segmental content at varying temporal scales using SSL mechanisms.


Given that the functions of prosody are tightly coupled with lexical content, it is unsurprising that representations of the full speech signal surpass MPM for tasks like emotion recognition. However, MPM is competitive for phrasal boundary and prominence detection, underscoring the value of prosody for these tasks.   
In the future, we hope to investigate the relative importance of pitch and energy features in this model, and more explicit inclusion of duration information. MPM also provides a means to compare prosodic systematicity across speech styles.

\newpage
\bibliographystyle{IEEEtran}
\bibliography{mybib}

\end{document}